\documentclass[sigconf]{acmart}

\usepackage{graphicx}
\usepackage{float}
\usepackage{subcaption}
\usepackage{listings}
\usepackage{geometry}
\usepackage{amsfonts}
\usepackage{multirow}
\usepackage{xcolor}
\usepackage{paralist}

\usepackage[disable]{todonotes}
\usepackage{url}

\usepackage{amsmath}

\usepackage[capitalise,nameinlink]{cleveref}

\newtheorem{definition}{Definition}

\copyrightyear{2022}
\acmYear{2022}
\setcopyright{rightsretained}
\acmConference[WWW '22 Companion]{Companion Proceedings of the Web Conference 2022}{April 25--29, 2022}{Virtual Event, Lyon, France}
\acmBooktitle{Companion Proceedings of the Web Conference 2022 (WWW '22 Companion), April 25--29, 2022, Virtual Event, Lyon, France}
\acmDOI{10.1145/3487553.3524719}
\acmISBN{978-1-4503-9130-6/22/04}

\begin{document}

\lefthyphenmin=4
\righthyphenmin=4
\sloppy
\title{Scaling R-GCN Training with Graph Summarization}

\author{Alessandro Generale}
\authornote{This paper is the outcome of the thesis work of the first author}
\affiliation{%
  \institution{Vrije Universiteit Amsterdam}
   \country{The Netherlands}
}

\author{Till Blume}
\affiliation{%
  \institution{Ernst \& Young GmbH WPG – R\&D}
  \city{Berlin}
  \country{Germany}}
\email{till.blume@de.ey.com}
\orcid{0000-0001-6970-9489}

\author{Michael Cochez}
\affiliation{%
  \institution{Vrije Universiteit Amsterdam}
  \institution{Elsevier Discovery Lab}
   \city{Amsterdam}
   \country{The Netherlands}
}
\email{m.cochez@vu.nl}
\orcid{0000-0001-5726-4638}

\begin{abstract}
Training of Relational Graph Convolutional Networks (R-GCN) is a memory intense task.
The amount of gradient information that needs to be stored during training for real-world graphs is often too large for the amount of memory available on most GPUs.
In this work, we experiment with the use of graph summarization techniques to compress the graph and hence reduce the amount of memory needed.
After training the R-GCN on the graph summary, we transfer the weights back to the original graph and attempt to perform inference on it.
We obtain reasonable results on the {\it AIFB}, {\it MUTAG} and {\it AM} datasets.	
Our experiments show that training on the graph summary can yield a comparable or higher accuracy to training on the original graphs.
Furthermore, if we take the time to compute the summary out of the equation, we observe that the smaller graph representations obtained with graph summarization methods reduces the computational overhead.
However, further experiments are needed to evaluate additional graph summary models and whether our findings also holds true for very large graphs.

\end{abstract}



\settopmatter{printfolios=true}
\maketitle

\todo[inline,color=lime]{8+2 references+2appendix}

\section{Introduction}
Knowledge Graphs (KG) emerged as an abstraction to represent and exploit complex data, and to ease accessibility~\cite{KG_definition}. 
As a result, extensive collections of data stored in KGs are now publicly available, spurring the interest in investigating novel technologies that aim to structure and analyze such data.
However, KGs, including well-known ones such as DBpedia and WikiData~\cite{linked_data}, remain incomplete.
There is an evident trade-off between the quantity of data available and its adequate coverage (completeness)~\cite{bloem2021kgbench}. %
Predicting missing information in KGs, e.g., predicting missing links, is the main focus of statistical relational learning~\cite{schlichtkrull2018}. 
However, such methods are challenged by large quantities of data and the unknown structure of KGs, harming the scalability of applications~\cite{10.1007/978-3-319-46547-0_20}.

Several techniques have been developed to tackle these larger graphs while still retaining their structure.
First, training of GCNs can be scaled by developing a memory-optimized implementation or distributing the computations over multiple GPUs~\cite{vasimuddin2021distGNN}. 
Second, some techniques try to work with condensed representation of the KG that retains the original structure of the graph.
For example, Salha et al.~\cite{salha2019degeneracy} proposed to use a highly dense subset of nodes from the original graph.
Deng and his co-authors~\cite{deng2020graphzoom} suggest the creation of a fused graph that embeds the topology of the original graph and in turn recursively coarsens into smaller graphs for a number of iterations.
The methods have been shown to improve classification accuracy and accelerate the graph embedding process~\cite{deng2020graphzoom}.

In this work, we propose to use graph summarization techniques to create a condensed representation of the KG, i.\,e., the graph summary.
In our approach, we train an R-GCN on the graph summary and, subsequently, transfer the obtained weights to an R-GCN based on the original KG.
Finally, we evaluate the later R-GCN and then investigate how its performance behaves when trained further, compared to an R-GCN  based on the full KG that was not pre-trained on a summary.
From our experiments we observe that training on the graph summary can yield a comparable or higher accuracy to training on the original graph. 
Furthermore, we show that the smaller graph representations obtained with graph summarization methods reduces the computational overhead if we do not incorporate the time needed for the summarization.

\section{Relational Graph Convolutional Networks}
Graph Convolutional Networks (GCNs) have spurred great interest in Machine Learning with graph entities.
A GCN takes as input a graph, potentially with input features for the nodes, and outputs embeddings for each of the nodes.
GCNs use a message-passing algorithm which means that neighboring information influences the embedding representation of a node.
However, a drawback of a GCN is that it treats all relations within a heterogeneous graph the same.
Hence, the R-GCN was proposed, which enables the inclusion of relational information into the Graph Neural Network
(originally proposed by Schlichtkrull et al.~\cite{schlichtkrull2018} and extensively analyzed by Thanapalasingam et al.\cite{thanapalasingam2021relational}).

An R-GCN is defined as a convolution that performs message passing in the context of multi-relational graphs.
The updated state of a node $i$ after message passing step $l+1$ of this graph neural network is as follows:
\begin{equation}
	\label{eq:RGCN}
	h_{i}^{(l+1)} = \sigma\left( \sum_{r \in R} \sum_{j \in N_{i}^{r}} \frac{1}{c_{i,r}} W_{r}^{(l)} h_{j}^{(l)} + W_{0}^{(l)}  h_{i}^{(l)} \right)
\end{equation}

This state is a combination of the state of the neighboring nodes after the previous message passing step $l$ and the previous state of the node itself.
In the formula, we sum over all relation types $r$ in $R$ and then over all neighbors $j$ which have a relation $r$ to node $i$ ($N_{i}^{r}$).
The state of these nodes is projected using relation specific matrices \(W_r^{(l)}\) (the weight matrix for the relation $r$ at layer $l$) before being summed.
The R-GCN adds a self-connection when updating the representation of a node, and learns a dedicated weight matrix \(W_0^{(l)}\) to project  nodes onto themselves.
Finally, the state is set to the value of that sum after applying a non-linear activation function.
The state after several message passing steps form a representation for each node in the graph. 
These representations can be used for node classification or other downstream machine learning tasks.

The R-GCN model suffers from over-parametrisation. %
Schlichtkrull et al.~\cite{schlichtkrull2018} use basis decomposition to reduce the number of weight matrices. %
With matrix decomposition, each of the relations has to learn %
relation coefficients to scale the single projection matrix. %
Alternatively, a block diagonal decomposition can be used to reduce the parameters~\cite{schlichtkrull2018}. %

\section{Graph summarization}
\label{sec:graph-summarization}

Representing data as a graph is increasingly popular since graphs allow a more efficient and a more flexible implementation of certain applications compared to relational databases~\cite{DBLP:journals/corr/abs-2003-02320,DBLP:conf/data/FernandesB18}.
Relational databases operate on tabular data models and employ strict data schema~\cite{harrington2016relational}.
Graph databases use labeled nodes to represent entities such as people or places and labeled edges to represent relationships between entities~\cite{DBLP:journals/corr/abs-2003-02320}.
The combination of node labels and edge labels can be considered the schema of the graph database~\cite{database_schema}.

In this paper, we define our (Knowledge) Graph as follows, and note that this definition includes RDF graphs\footnote{\url{https://www.w3.org/TR/rdf11-concepts/}} as a special case.

\begin{definition}[Graph](definition obtained from~\cite{campinas2016graph} page 15). \newline	 
\noindent Let \(V\) be a set of nodes and \(L\) a set of labels.
The set of directed labeled edges is defined as \(A\subseteq \{(x, \alpha, y)~|~(x,y) \in V^2, \alpha \in L\}\). A Graph \(G\) is a tuple \(G = (V, A, l_V)\) where \(l_V : V \Rightarrow L\) is a node-labeling function.
\label{sec:def_1}
\end{definition}
In the context of RDF web graphs, it is common practice to think of the graph as a set of triples $(x, \alpha, y)$. 
Such triple has a head or source node \(x\) that has a direct relationship  \(\alpha\) to a destination node \(y\), also referred to as target or object node.

\begin{figure}
	\centering
	\includegraphics[width=\linewidth]{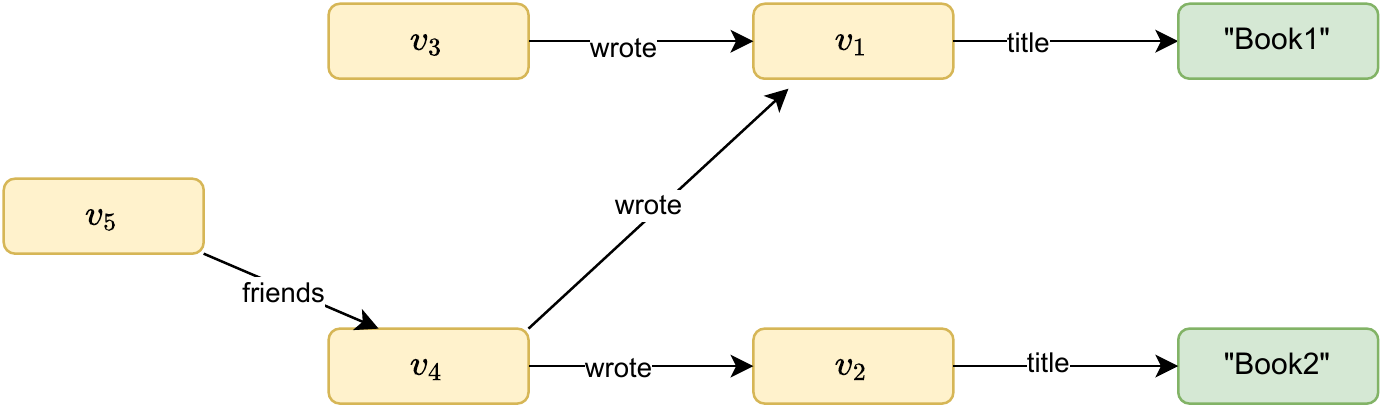}
	\caption{Fragment of a KG.
		The entities are the nodes \(v_1\) to \(v_5\) and the edges have attributes e.g.
		'wrote', 'title' and 'friends'. The green nodes are Literals.}
	\label{fig:kg_fragment}
\end{figure}

\Cref{fig:kg_fragment} displays a small example KG. %
The set of nodes \(\{v_1, v_2, v_3, v_4, v_5\}\) are entity nodes.
The set of nodes \(\{"Book1", "Book2"\}\) are literal nodes, normally used for values such a strings, numbers or dates.
An example of triple within this graph is the following: \((v_1, \text{title}, \text{"Book1"})\). 
Following~\cite{campinas2016graph}, we define an attribute to be the label of an edge.
In this case the attribute relating node \(v_1\) and literal node "Book1" is \emph{title}.

Such Knowledge Graphs allow to flexibly add new node labels and edge labels, i.\,e., adapt to new entities and relationships~\cite{KG_definition}.
This flexibility regarding the data schema means that the inherent structure of the data may be unclear at first or evolve during the lifetime of a graph database. 
Graph summarization facilitates the identification of meaning and structure in data graphs~\cite{DBLP:journals/csur/LiuSDK18,Goasdou2020RDFGS}.
Thus, graph summaries can be used, among other applications, to discover the data schema of an existing graph database~\cite{DBLP:journals/vldb/CebiricGKKMTZ19}.
When we describe a graph summarization techniques, we typically characterize it with three dimensions~\cite{campinas2016graph}:
\begin{inparaenum}[(I)]
    \item size of the graph,
    \item size of the graph summary,
    \item the impact of the graph’s heterogeneity on the graph summary.
\end{inparaenum}

When we refer to the size of a graph we generally define it as the number of edges \(|A|\) that the graph is composed of~\cite{campinas2016graph}. 
A fourth dimension of graph summarization is the role of the literal nodes.
Often, summaries are computed taking into consideration literal nodes but their content is abstracted away.
As a result, the compression rate increases because many nodes within the graph are ignored when computing the summary.
For instance, looking back at \Cref{fig:kg_fragment}, the triples connecting to literal nodes "Book1" and "Book2" are stripped away and then the summary is computed.

Intuitively, graph summaries are condensed representations of graphs such that a set of chosen features of the graph summary are equivalent to the features in the original graph~\cite{DBLP:journals/tcs/BlumeRS21}.
To achieve this, structural graph summaries partition nodes based on equivalent subgraphs.
To determine equivalent subgraphs, features such as specific combinations of labels in that subgraph are used.
We call this the schema of nodes. 

A graph is considered heterogeneous when not consistent regarding its labeling and structure, e.\,g., there is a huge variety in schema used to describe entities and their relationship~\cite{campinas2016graph}. 
The more heterogeneous a graph, the more complex it becomes to create a concise and precise summary.
Concise refers to the size of the summary, whereas precise concerns the preservation of the entity structure.
In practice, a reasonable summary is typically smaller than the original graph~\cite{DBLP:journals/rpjdi/ScherpB21}.
Below, we discuss two families of graph summaries: approximate graph summaries and precise graph summaries.

\subsection{Approximate Graph Summaries}
Approximate graph summaries exploit the different features of the data that form the local information of a node.
For instance, a node has incoming and outgoing edges with different labels also called attributes~\cite{campinas2016graph}. The direction of an edge has semantics within the structure of the graph.
A node has a one-to-many \texttt{rdf:type} relations to an object node defining a specific property of a node.
The local information of a node described above is combined to produce different partitions of the original graph.
The combination of information defines the summarization relation.
The partitions, which correspond to summary nodes, are produced based on the summarization relation.
Nodes belonging to the same partition are aggregated and mapped accordingly.
This section of the paper discusses several approximate graph summary methods such as Attributes Summary, IO Summary and Incoming Attributes summary.

\subsubsection{Attributes Summary}
\label{sec:attribute-summary}
\begin{figure}[H]
    \centering
    \includegraphics[width=\linewidth]{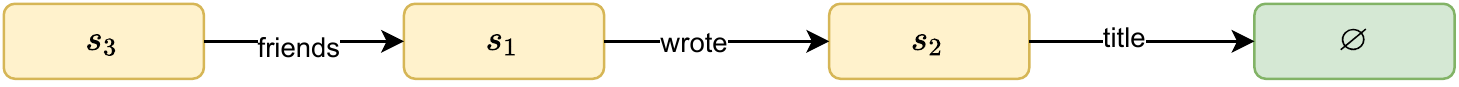}
    \caption{Attributes Summary graph with generated mapping from original node to summary node.
The green node labeled by the empty set symbol \(\empty\) abstracts the content of Literal nodes.}
    \label{fig:fig_summaryAtt}
\end{figure}

The Attributes Summary has been defined in previous literature~\cite{campinas2016graph,Blume_2021} as an approximate graph summary.
In the context of a KG, each entity node has zero-to-many relations with different attributes.
The set of the attribute per nodes is computed and the entities that have equivalent attribute sets are partitioned together.
Each partition is then considered a unique summary node.
Each of the original entity nodes is mapped to a partition, hence a summary node, if it has at least one outgoing edge and is not a literal node.
It is important to notice the summary method being discussed aggregates a node with few edges into the same partition as another node with many more edges if they have equivalent attribute sets.
Therefore, the number of relations a node has is not considered.
The nodes that do not have any outgoing edges, often the literal nodes in KGs, are aggregated to the same partition or summary node.
The literal nodes information when performing the following summarization technique is abstracted away.
\Cref{fig:fig_summaryAtt} shows the Attributes Summary result on the KG fragment presented in \Cref{fig:kg_fragment}. 

\begin{definition}[Summarization Relation \(R_a\)](definition obtained from~\cite{campinas2016graph} page 48). \newline 
    \noindent Let \(G = (V, A, l_V)\) and \(S_a = (W_a, B_a, l_{W_{a}})\) be two graphs.
G refers to the original graph.
\(S_a\) refers to the summary graph.
The set of nodes \(W_a\) contains as many elements as the power set of attributes in the graph.
Therefore, \(S_a\) corresponds to the Attribute Summary of G according to the summarization relation \(R_a \subseteq V \times W_a\) defined as follows:
    \newline \center{\(R_a = \{(u,x) \in V * V_a~|~attributes(u) = attributes(x)\}\)}
\label{sec:def_Ra}
\end{definition}
\noindent This means the nodes are placed in the same partition if the sets containing their attributes are the same.
Note that no duplicates triples are added as we store the triples into a set, and by definition a set does not contain duplicate items.

Computing this relation and assigning a new label to each partition, we obtain the mapping in~\Cref{tab:mapping_Ra}.
We can no use this mapping to summarize our running example from \Cref{fig:kg_fragment} into the graph in \Cref{fig:fig_summaryAtt}.
For example, the nodes \(v_1\) and \(v_2\) belong to the same partition as they share the same attribute set \(\{title\}\) and hence they are summarized into the same node, which received the label \(s_2\).

The summarization relation described above maps nodes without any outgoing edges to an \emph{empty summary node}, which we give the label \(\emptyset\).
Usually that node represents literal nodes as they do not have any outgoing edges.
In our example, the string literals `Book1' and `Book2' map to the same empty summary node.

\vspace*{0.5cm}

\begin{table}[H]
\center
\begin{tabular}{cc} \toprule
	{$ V $} & {$ W_{a} $}                    \\ \midrule
	$ v_{1} $, $ v_{2}$   & $ s_{2} $    \\
	$ v_{3}$, $ v_{4} $   & $ s_{1} $    \\
	$ v_{5} $  & $ s_{3} $              \\
\bottomrule
\end{tabular}
\caption{ The table indicates the mapping produced by the Attributes Summary from original nodes \(\in V\) to summary nodes \(\in W_a\) under \(R_a(V, W_a)\).}
\label{tab:mapping_Ra}
\end{table}

\subsubsection{IO Summary}
The Input-Output summary is similar to the Attributes Summary.
In fact, the structure of \Cref{sec:def_Rio} is almost identical.
However, in this case, for each node in the graph it creates two sets defined as incoming attributes set and outgoing attributes set.
All the nodes that have equivalent incoming and outgoing attributes sets are inserted in the same partition.

\begin{definition}[Summarization Relation \(R_{io}\)] (definition obtained from~\cite{campinas2016graph} page 48-49). \newline 
    \noindent Let \(G = (V, A, l_V)\) and \(S_{io} = (W_{io}, B_{io}, l_{W_{io}})\) be two graphs.
G refers to the original graph.
\(S_{io}\) refers to the summary graph.
The set of nodes \(W_{io}\) contains as many elements as the power set of attributes in the graph.
Therefore, \(S_a\) corresponds to the Incoming Outgoing Attribute Summary of G according to the summarization relation \(R_{io} \subseteq V \times W_{io}\) defined as follows:
    \newline \center{\(R_{io} = \{(u,x) \in V * V_{io}~|~attributes(u) = attributes(x) \wedge attributes^{-1}(u) =
    attributes^{-1}(x)\}\)}
\label{sec:def_Rio}
\end{definition}
The function \(attribute\) and \(attribute^{-1}\) compute the outgoing edges partition and the incoming edges partition respectively.
The negative exponent indicates an inverse operation.
Therefore, instead of computing the attributes sets of outgoing edges, it computes the attributes sets of incoming edges.
A node \(u\) that has a equivalent incoming and outgoing attributes sets to a node \(x\) shows that node \(u\) and \(x\) belong to the partition under the summarization relation \(R_{io}\). 

\subsubsection{Incoming Attributes Summary}
The Incoming Attributes Summary is the inverse operation of the Attributes Summary.
The following partitions each node based on the incoming attributes set.
Two nodes that have equivalent incoming attributes sets are inserted into the same partition.
\Cref{sec:def_Ria} has a similar structure to the other previously mentioned summarization techniques.

\begin{definition}[Summarization Relation \(R_{ia}\)](definition obtained from~\cite{campinas2016graph} page 50). \newline 
    \noindent
    Let \(G = (V, A, l_V)\) and \(S_{ia} = (W_{ia}, B_{ia}, l_{W_{ia}})\) be two graphs.
G refers to the original graph.
\(S_{ia}\) refers to the summary graph.
The set of nodes \(W_{ia}\) contains as many elements as the power set of attributes in the graph.
Therefore, \(S_{ia}\) corresponds to the Incoming Attributes Summary of G according to the summarization relation \(R_{ia} \subseteq V \times W_a\) defined as follows:
    \newline \center{\(R_{ia} = \{(u,x) \in V * V_{ia}~|~attributes^{-1}(u) = attributes^{-1}(x)\}\)}
\label{sec:def_Ria}
\end{definition}
The function \(attribute^{-1}\) takes as input the original node and outputs the corresponding summary node based on the partitions formed on the incoming edges of a node.
Therefore, a node \(u\) with an equivalent incoming edges partition to node \(x\) shows that nodes \(u\) and \(x\) belong to the same summarization relation \(R_{ia}\).

\subsection{Precise Graph Summaries}
Partitioning nodes based solely based on their local subgraph schema structures has certain drawbacks.
In contrast to such approximate graph summaries, precise graph summaries compute the schema of nodes considering neighboring nodes over multiple hops~\cite{Kaushik2002ExploitingLS,chen2003dkindex}. 
Similarly, the notion of aggregating nodes based on local and neighboring information is shared in modern machine learning models that deal with graph entities.
The general purpose of a graph summary is to mirror the structure of the original graph whilst being consistently smaller in size.
Precise graph summaries are considered a stricter method because the summarization relation is more complex and takes more information from the original graph into account.
In order for two nodes to belong to the same partition, neighboring nodes must also share the same schema.
A graph summary is said to be precise when indiscernible from the original graph~\cite{campinas2016graph}. 
A summary graph may not have paths that are present in the entity graph but the overall structure of the entity graph is preserved due to graph homomorphism~\cite{campinas2016graph}. 
Additionally, the structure of the original graph is kept in the summary but the inverse may not hold true.
Comparing the mapping in \Cref{tab:mapping_Ra}, there is no unique way to always reconstruct the original KG.

\begin{definition}[Precise Graph Summary](definition obtained from~\cite{campinas2016graph} page 41). \newline
\noindent Let \(G = (V, E, R)\) and \(S = (W, B, L_W)\) be two graphs.
The graph S is the summary of the graph G according to a summarization relation \(R \subseteq V \times W\). Let \(p = (x_1, \alpha_1, x_2) \in B \wedge ~ .. \wedge ~ (x_n, \alpha_n, x_{n+1}) \in B\) be a path in the summary graph S where \((x_1, .., x_{n+1}) \in W^{n+1} \). Let the set \(\{u_1, .., u_{n+1}\}\) be a summary path instance \(u_i \in V\). A summary is therefore called precise when each instance of a summary path in p forms a path in the original entity graph with regards to the edges in~p.

    \center{\(\forall \in [1, n] ~\exists ~(u_i, \alpha_i, u_{n+1}) \in E\)}
\label{sec:def_5}
\end{definition}
\Cref{sec:def_5} states that a graph summary is fully precise if all paths that can be formed in the summary graph can be found in the original graph.
The original graph may hold paths that are not present in the graph summary.
Looking at the example of graph summary produced in \Cref{fig:fig_summaryAtt}, there exist paths which are not present in the original graph, such as the paths \((v_5, v_3, v_2)\). Such path can be formed in the summary graph as the nodes \(\{v_5, v_3, v_2\}\) are partitioned by the nodes \(\{s_3, s_1, s_2\}\) which form the previously mentioned path.
With the same subset of summary nodes, it is possible to create paths that are contained in the original graph, for instance, the path \((v_5, v_4, v_2)\). 

\subsubsection{Bisimulation}
The intuition behind bisimulation, in the context of graphs, is the interpretation of the graph data as transition systems to discover structurally equivalent parts~\cite{Blume_2021,10.1145/1516507.1516510}. 
Two nodes' states are bisimilar if their states change following equivalent edge relations~\cite{Blume_2021}. 
Bisimulation is a binary relation that relates two arbitrary nodes in a KG when itself and its inverse are simulations~\cite{campinas2016graph}. 
For instance, consider two nodes {\it u} and {\it v}, a simulation relation states that an edge with type \(\alpha\) departing from \(u\) and pointing to an arbitrary node \(x\) implies that there exists an edge with type \(\alpha\) from \(v\) pointing to an arbitrary node \(y\) such that \(x\) and \(y\) are simulations.
It is useful to realize that two nodes are bisimilar if they share the same outgoing paths.
The notion of bisimulation is stricter as it is a symmetric equivalence relation, ensuring that each node can substitute one another.

\subsubsection{(k)-forward Bisimulation}
\label{sec:forward-bisim}
\begin{figure}
        \begin{subfigure}{\linewidth}
                \centering
                \includegraphics[width=0.75\linewidth]{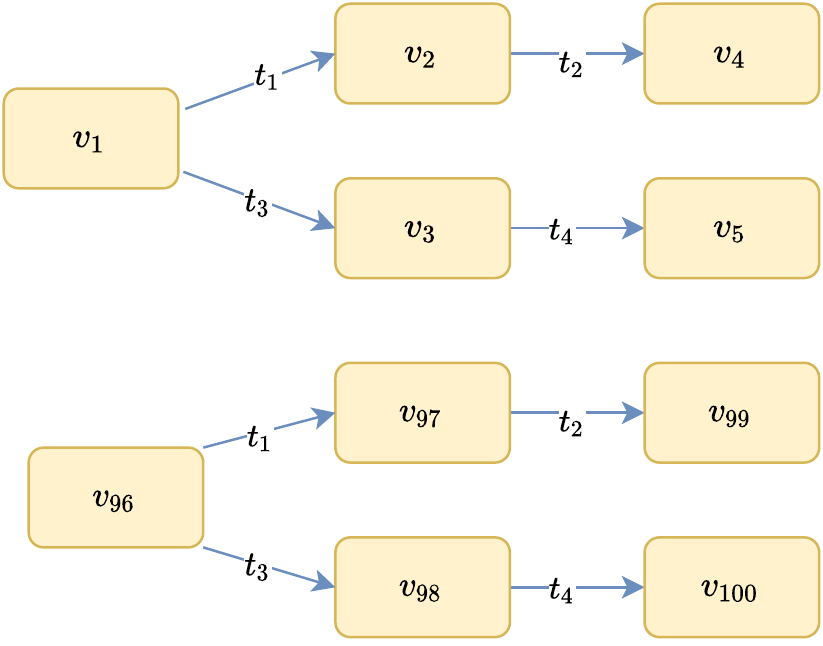}
                \caption {Fraction of a KG}
                \label{fig:bisim_orig_a}
        \end{subfigure}%
    \\
        \begin{subfigure}{\linewidth}
                \centering
                \includegraphics[width=0.8\linewidth]{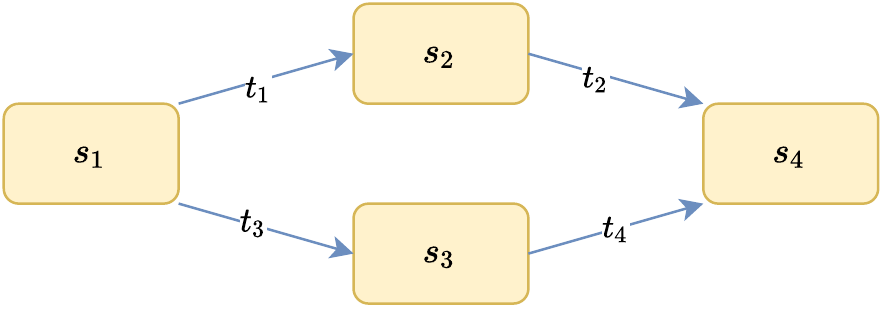}
                \caption {(k)-bisimulation Graph Summary}
                \label{fig:bisim_orig_sum_b}
        \end{subfigure}%
        \caption{\Cref{fig:bisim_orig_a} shows the original KG fragment.
\Cref{fig:bisim_orig_sum_b} shows (k)-bisimulation graph summary produced with a chaining parameter \(k=3\).}
\end{figure}

(k)-forward bisimulation is an example of the many graph summarization techniques which aggregates nodes based on neighbors' schema over multiple hops~\cite{Kaushik2002ExploitingLS,KONRATH201252,%
	6226408}. 
It has also been defined as a stratified bisimulation in~\cite{Blume_2021}, a summarization relation that is restricted to a maximum path length of k-edges.
Other summarization techniques that propose bisimulation over {\it k} hops also consider the edge direction~\cite{10.1145/1516507.1516510}. 
These are referred to as {\it forward}, {\it backward} and {\it forward/backward}~\cite{kaushik2002covering} bisimulation.
In this paper, we use the {\it FLUID} framework~\cite{blume2020incremental}, which allows us to create a {\it forward} (k)-bisimulation summary.
\Cref{fig:bisim_orig_a} represents a small fragment of a KG.
A forward (k)-bisimulation with chaining parameter \(k=3\) is performed on this small entity graph. %
The resulting summary graph with alongside the one-to-one mapping from original node to summary node is shown in \Cref{fig:bisim_orig_sum_b}.

\section{Methods}
In this work, we focus on multi-label classification on the types of the nodes.
Therefore, the R-GCN uses of a summarized version of the original entity graph and infers the \texttt{rdf:type} relations for each of the original nodes.
We expect that the summarized version of the graph retains enough structure of the entity graph, helping to generalize to the complex hierarchy of types in different nodes.
It is possible to transfer the parameters of the summarized nodes back to the original nodes by storing links between nodes in the original graph and in the graph summary.
This allows us to evaluate the performance of a model only trained on the graph summary and, possibly, to continue training on the original graph.
As baseline, we train an R-GCN only on the original graph.
All used R-GCN models follow the architecture proposed in~\cite{schlichtkrull2018}. 
However, the models in our work use {\it binary cross-entropy} loss and apply a sigmoid activation function on the output to predict multiple labels.

An issue raised by Bloem et al.~\cite{bloem2021kgbench} is the low number of labeled nodes in the datasets.
This is an evident problem when it comes to testing the actual performance.
To counter this issue we capture a specific relation that is present in most nodes and use its value as a label.
We use the \texttt{rdf:type} relation which indicates the type of the node (a node can have multiple types) and use these as the classes for the nodes.
This is explained in more detail in \Cref{sec:summaryLabels}. 
The R-GCN models used in this work are implemented using Python 3.9.0 with the PyTorch geometric~\cite{Fey/Lenssen/2019} framework.

\subsection{Data Pre-processing \& Summary Generation}
The first step of the experiment is to create a map of original nodes to their labels.
There are several ways to obtain these labels.
It might be that they are not present in the graph, but rather retrieved from an external source.
Alternatively, they are obtained from a set of chosen properties available in the graph~\cite{8970828}.
In both cases nodes may belong to a single or multiple classes.

However, there are only few datasets equipped with a reasonable number of labeled nodes~\cite{bloem2021kgbench}. 
Hence, for our experiments we obtain labels by considering the \texttt{rdf:type} relation, and removing these triples from the original graph.
Other properties of nodes within the KG are used to train the model.

The graph that was stripped of the \texttt{rdf:type} triples is used as an input for the two summarization frameworks to compute the Attributes Summary (see \cref{sec:attribute-summary}) and the forward (k)-bisimulation (see \cref{sec:forward-bisim}) with a chaining parameter \(k = 3\). 
To compute the Attribute Summary, we use a SPARQL query developed by Campinas~\cite{campinas2016graph} and a slightly modified version which also gives us the mappings.
To compute the forward (k)-bisimulation, we use the FLUID framework~\cite{DBLP:journals/tcs/BlumeRS21,blume2020incremental}.
We then also generate the mapping from the summary nodes to the original nodes.

Then, we create the summary graph by iterating over the edges in the original graph.
For each edge, we apply the mapping function on the source and destination node, and use the outcomes and the relation type to form a new edge for our summarized graph.
At this step, we also filter out the literal nodes and remove triples containing the Web Ontology Language attributes as suggested by Campinas~\cite{campinas2016graph}.

\subsection{Summary Nodes Labels}
\label{sec:summaryLabels}
A summarization technique may map multiple nodes with different \texttt{rdf:type} values to the same partition or summary node.
And hence, we have multiple labels for the same summary node.
We create a \emph{weighted multi-label} for each summary node by taking the relative frequency of each type in the partition (i.e., all nodes mapping to the summary node).

\begin{figure*}	
\centering
   \includegraphics[width=14cm]{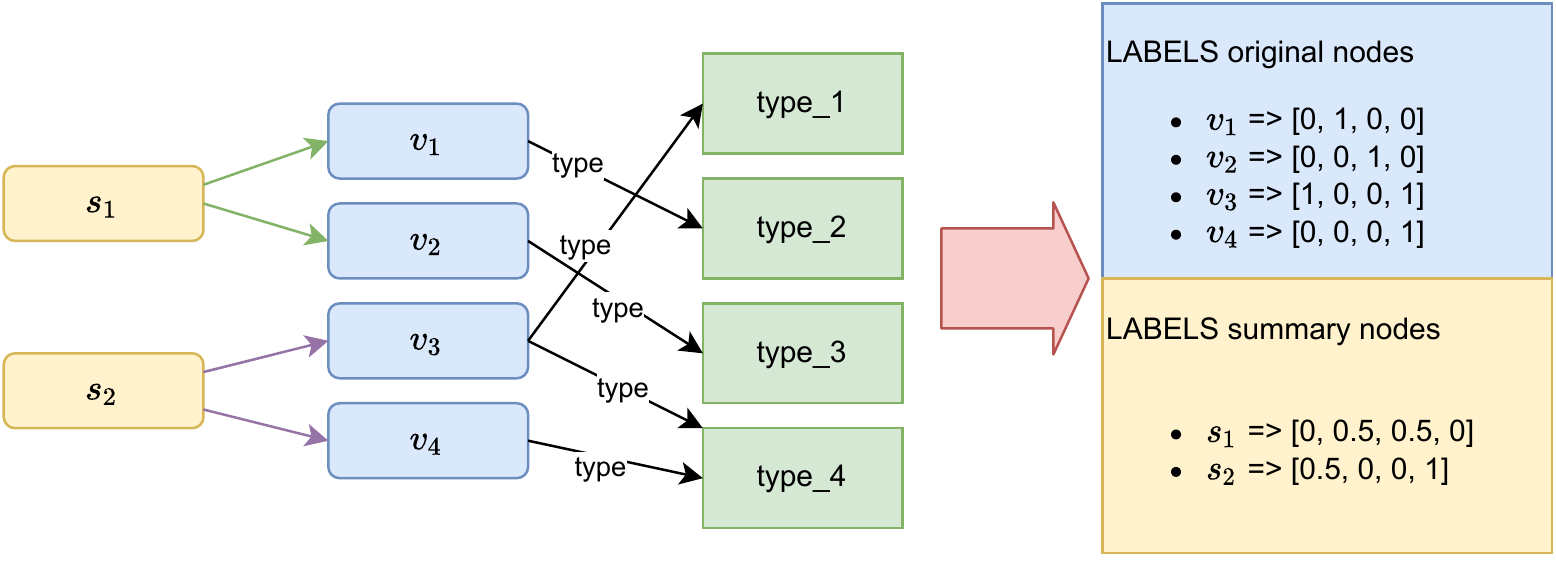}
   \caption{Diagram showing the original and summary nodes labeling process.}
   \label{fig:labels}
\end{figure*}

\Cref{fig:labels} shows an example of the labeling of the summary and original nodes.
The nodes \(s_1\) and \(s_2\) are two summary nodes that represents original nodes \(\{v_1, v_2\}\) and \(\{v_3, v_4\}\) respectively.
The node \({v_1}\) has a \texttt{rdf:type} relation and is labeled \(type_2\). Similarly, \(v_3\) has two \texttt{rdf:type} relations and is therefore labeled \(type_1\) and \(type_4\). From the partitioning of the summary nodes we can obtain the labels for the summary nodes.
The node \(s_2\) represents nodes \(\{v_3, v_4\}\) and its labeling is 0.5 for \(type_1\) because only one node has that relation.
The \(type_4\) is labeled 1.0 because both nodes point to \(type_4\). 

The machine learning model that is trained with the summary representation uses these weighted labels.
The transfer learning model and benchmark model (our baseline) train on binary values to predict the types of nodes.

\subsection{Machine Learning Model: R-GCN}
To perform the experiments, we set up three different R-GCN models, a model that learns on the graph summary, a second model with parameters transferred from the graph summary model, and finally, a model that follows a normal initialization to act as a baseline.
The initialization of the second model is performed by transferring the embedded parameters from the model that learns on the graph summary using the mapping from original nodes to summary nodes.

As mentioned earlier, the model has been slightly modified in order to fit the purpose of the multi-label classification task.
A {\it binary cross-entropy loss} function~\cite{NEURIPS2019_9015} is used to compute the loss during training.
The assumption is that an element that belongs to one class does not influence the probability of the same element belonging to another class.
The following loss function requires target values to be between 0 and 1. 
The target values of the summary and original nodes are created as described in \cref{sec:summaryLabels} %
The labeling process gives values to the labels that depending on the frequency of that label applying to the original nodes inside the partition.
A sigmoid activation function restricts the output of all models between 0 and 1, corresponding to the probabilities of the labels.
The output of the model is then rounded to the nearest integer to match the original node labels.

A parameter that needs to be taken into consideration is the number of hidden units between the two R-GCN convolutions.
Following the literature~\cite{kipf2017semisupervised,schlichtkrull2018}, we set the number of hidden values to 16 for all used R-GCN models.
As proposed in~\cite{kipf2017semisupervised}, we also do not employ a regularizer.
The Adam optimizer~\cite{NEURIPS2019_9015} with a learning rate of \(1.0 * 10^{-2}\) and a weight decay of \(5.0 * 10^{-4}\) is applied as suggested in~\cite{schlichtkrull2018}. 
We do not use basis decomposition.

The model that learns on the summary graph is trained first for a total of 51 epochs.
The parameters of this model are transferred to the second R-GCN through the mapping produced by the summarization technique.
An additional training parameter can be set which consists of freezing the layers of one or more layers so that the weights do not update during training.
This is suggested to be done on the model to which the weights have been transferred to run a faster training.
If the first convolutional layer of the transfer learning model is frozen, the embedded parameters that were transferred from the graph summary are not updated in the backward pass.
The parameters of the R-GCNs are detailed in \Cref{sec:parameters}. 
The labeled instances are split 80\% for training and 20\% for testing.

\begin{figure}[!h]
\centering
   \includegraphics[width=\linewidth]{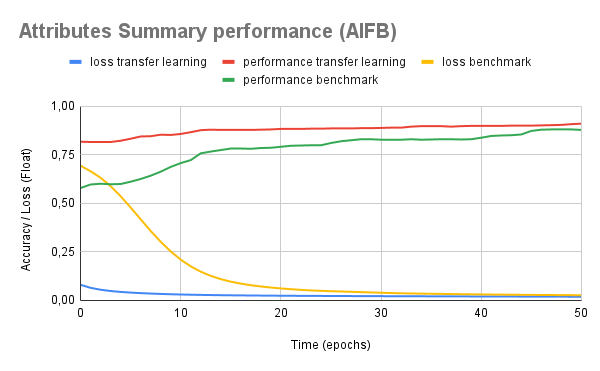}
   \caption{Graph showing the performance of the Attributes Summary performed on the dataset AIFB. The transfer learning model (red line) starts predicting at a higher accuracy than the baseline (green line) from early iterations and converges to a better optimal solution.}
   \label{fig:performance_attsumm_AIFB} 
\end{figure}

\section{Data sets}
\label{sec:datasets}

Following existing research works~\cite{schlichtkrull2018,bloem2021kgbench}, we conduct our experiments on the AIFB, MUTAG and AM datasets. 
Statistics of these datasets can be found in \Cref{table:datasets}.

\begin{table}[H]
	\center
	\caption{Statistics of commonly used RDF format datasets in research}
	\label{table:datasets}
	\begin{tabular}{l|rrrr}
		\toprule
		& \textbf{AIFB} & \textbf{AIFB} & \textbf{BGS} & \textbf{AM} \\ 
		\textbf{Entities}  & 8,285          & 23,644        & 333,845      & 1,666,764   \\ 
		\textbf{Relations} & 45            & 23            & 103          & 133         \\ 
		\textbf{Edges}     & 29,043        & 74,227        & 916,199      & 5,988,321   \\ 
		\textbf{Classes}   & 26             & 113             & 1            & 21            \\ 
		\bottomrule
	\end{tabular}

\end{table}

\section{Experimental Results}

\begin{table*}
	\centering
	\begin{tabular}{|c|c|cc|cc|cc|c|}
		\hline
		\multicolumn{2}{|c|}{} &\multicolumn{2}{|c|}{Attributes Summary} &\multicolumn{2}{|c|}{(3)-forward bisimulation} &\multicolumn{2}{|c|}{Baseline}  \\
		\cline{3-8}
		\multicolumn{2}{|c|}{}  & No training & After 50 epochs & No training & After 50 epochs & No training & After 50 epochs \\\hline
		\multirow{3}{*}{\rotatebox{90}{Dataset }}  & AIFB & \(82.52 \pm 3.93\) &  \(92.54 \pm 2.42 \)&  \(77.53 \pm 4.72\) &  \(91.30 \pm 2.03\) & \(58.03 \pm 14.15\) & \(87.98 \pm 2.13\) \\\cline{2-8}    &MUTAG & \(0.21 \pm 0.41\) &  \(35.07 \pm 9.70\) & \(15.94 \pm 13.47\) &\(36.56 \pm 8.56\) & \(24.77 \pm 16.33\) & \(28.77 \pm 2.00\) \\\cline{2-8}
		&AM & \(62.36 \pm 3.61\) &  \(80.14 \pm 2.17\) &  \(61.75 \pm 3.83\) &  \(72.63 \pm 5.31\) & \(13.90 \pm 11.02\) & \(78.63 \pm 3.21\) \\\cline{2-8}
		\hline 
	\end{tabular}
	\caption{Results of the experiments with (k)-fold cross validation with k set to 5. Note that this cross validation was not performed for the AM dataset due to computational cost; only two runs are reported for each model.
}
	\label{tbl:results}
\end{table*}

\todo[inline,color=lime]{MC I took a paragraph away here. It was just repeating from before, but less clear.}

We are interested in several results. First, we want to know how the transfer model, which gets initialized with the parameters from the summarized model performs.
Secondly, we want to know how its performance changes with respect to the epochs if we perform more training on this transfer model.
Finally, we have to compare these performances to the R-GCN which was no initialized had nothing to do with the summarized model (i.e., the baseline).
In our graphs, we show 2 performance curves. The first one is the transfer learning model. At the zeroth epoch, this shows the performance without any additional training. After that, it shows the performance with additional training.
The second curve shows the performance of the standard R-GCN with random initialization in function of the training epochs.
In our graphs, we also include the value of the respective loss functions.

\Cref{fig:performance_attsumm_AIFB} shows the performance of the transfer learning model trained on the Attributes Summary against the baseline for a single run of the on the {\it AIFB} dataset.
\Cref{tbl:results} shows the average starting and converging accuracy.
The average is calculated by performing (k)-fold cross validation with k set to 5. 
The red line represents the model to which the embedded parameters of the summary model are transferred to.
The green line represents the original model, with weights that are randomly initialized.
The transfer learning model (red line) starts with a reasonably high accuracy of 81.7\% for the first few epochs. 
The accuracy of the transfer learning model steadily increases up to 91.06\% after 50 epochs.
In contrast, the accuracy of the original model (green line) starts at a lower accuracy of 57.9\% and converges at a lower accuracy of 87.82\%.

\todo[inline,color=lime]{MC I have an issue here. I do not recall whether the curves are showing results for the validation or the test set. This might explain the difference. I took the explicit sentence about it below away.}

The results for the transfer learning model trained on the (k)-forward bisimulation can be found in \Cref{fig:aifb_kb}.
Summarily to the transfer learning model trained on the Attribute Summary, the transfer learning model trained on the (k)-forward bisimulation summary consistently outperforms the baseline model over all training epochs. 
This transfer learning model starts with an accuracy of 77.53\% and converges to a final accuracy of 91.30\% after 50 training epochs.

The dataset {\it AIFB} contains 8,285 entities, of which type information has been stripped.
The Attributes Summary reduces the size in total number of edges by 63.7\% on the {\it AIFB} KG.
In this experiment, we need fewer training epochs for the transfer learning model to predict with a reasonably good accuracy compared to the baseline model. 
Additionally, we observe that the transfer learning model consistently outperforms the baseline model when trained for the same number of epochs.

\todo[inline,color=lime]{A simple description of the results is missing, we jump directly into a discussion. I agree with MC that this analysis is also over critical. -- TBL}
\todo[inline,color=lime]{MC I am a bit confused with the first part of your analysis.
You look for issues with the summary model, while what we see is that the summarized model performs about the same as the baseline.
So it is rather the task/dataset and not something you have introduced.}
\todo[inline,color=lime]{MC I now rewrote parts to reflect the actually pretty nice results.}

The {\it MUTAG} dataset provided similar insights (see \Cref{sec:mutagPCKB} for single runs). 
However, the positive effect of the summarization is not that large. 
After 20 to 25 epochs, the baseline model catches up with the summarized model.
We think this difference has to do with the fact that both summarization techniques provided very concise summaries of the dataset.
This means that the model learned on tiny small versions of the original graph.
The compression rates were 93.1\% and 85.4\% for the Attributes Summary and (k)-forward bisimulation, respectively.
In addition to this, 113 different labels were found when producing the training and testing nodes in the data preprocessing step, making this an overall much harder task.
In previous work, node classification on the {\it MUTAG} dataset was done considering as few as 2 classes~\cite{thanapalasingam2021relational}.

Finally, we look at the largest KG, the {\it AM} dataset (see \Cref{sec:amPC} for single runs). 
We notice that the transfer models show better results to the baseline models at the start, but that after a while the performance of the baseline catches up to become roughly the same.
\todo[inline,color=lime]{MC I took out some details here. They did not add much and were over critical regarding non statistically significant differences in performance.}

The graph summaries which we used in our experiments consider the relations with literal nodes as well (see \cref{sec:graph-summarization}).
As an ablation, we conducted experiments where we do not take the literal nodes into account when computing the summary.
The results for single runs on the {\it AIFB} and {\it MUTAG} datasets can be found in \Cref{sec:aifbPCnoLit}. 
What we note is that the performance of the transfer learning models becomes less consistent and often fall below the baseline. 
Hence, considering literal information seems critical for the performance of transfer learning models trained on graph summaries. 

\section{Conclusion and Future Work}
We introduced an approach to use graph summaries to efficiently and effectively train R-GCN models on KGs.
In our experiments, the R-GCN models trained on the graph summary often outperformed the baseline models and typically started at a much higher initial accuracy before the first training epoch. 
The only exception is the model trained on the forward (k) bisimulation summary on the {\it MUTAG} dataset, which yields at 4\% lower accuracy after 50 training epochs than the baseline. 
In all other cases, the transfer learning model outperforms the baseline.
What we identify is that transferring the parameters learned from the graph summary model leads to a jump-start of the learning process.
This supports the importance and relevance of graph summarization methods, whose smaller graph representations scale down and reduce the computational overhead involved with novel machine learning models dealing with large KGs.
However, due to the complexity of the classification task, the variation in the accuracy remains large.
The summarization relation (k)-forward bisimulation and Attributes Summary behaved similarly in the training behavior for the {\it AIFB} and {\it AM} datasets.

\todo[inline,color=lime]{MC There was a limitation here which I think is not a limitation at all. We perform a multi-labeling tasks, so what?}

For future work, we plan to extend the experiment with additional tasks. 
It might be possible to include a baseline study comparing the results with previously reported performances on single-class prediction.
Besides, while the graphs we used are common in heterogeneous graph neural networks literature, experiments with larger graphs are necessary.

Another aspect to further experiment with is the hyperparameter k for the (k)-forward bisimulation.
In our experiments, the parameter k was fixed to 3. 
It remains open to investigate a possible relationship between the optimal hyperparameter k in the graph summary and the optimal number of layers in an R-GCN.

In addition to these, the model trained with the graph summary can only learn for the nodes that are mapped.
A number of nodes is discarded because such entities are not mapped when producing the graph summary, so their embedding will initially be random.
The number of discarded nodes can be reduced by increasing k or by using more complex summarization methods.

Finally, one could also investigate whether these summarization techniques can also be used for other graph neural networks besides R-GCN models, for example for the CompGCN \cite{vashishth2020compositionbased}. 

In this work, we established a framework that uses graph summaries to efficiently and efficiently train R-GCN to predict nodes' labels.
The model's performance observed and reported for the {\it AIFB} and {\it AM} datasets for both summarization methods suggests that summarization can be used to scale training without harming the performance and even increase the performance.

\todo[inline,color=lime]{MC Future work CompGCN \cite{vashishth2020compositionbased}

This is an idea for future research to explore more complex summarization techniques.

Future work on other methods, which could also include more traditional link prediction methods.
}

\begin{acks}
We used the DAS5 system~\cite{das} for training the machine learning models. %
Part of this research was funded by Elsevier's discovery Lab.
\end{acks}

\bibliographystyle{ACM-Reference-Format}
\bibliography{main.bib}

\appendix

\section{Model Parameters}
\label{sec:parameters}
\begin{table}[H]
\center
\begin{tabular}{l|rrrr}
\toprule
                                & \textbf{Summary} & \textbf{TransferL} & \textbf{baseline} &  \\ 
                            
\textbf{number hidden units}    & 16               & 16                 & 16                 &  \\ 
\textbf{number R-GCN layers}     & 2                & 2                  & 2                  &  \\ 
\textbf{learning rate}          & 0.01             & 0.01               & 0.01               &  \\ 
\textbf{weight decay}           & 0.0005           & 0.0005             & 0.0005             &  \\ 
\textbf{number of bases}        & None             & None               & None               &  \\ 
\textbf{layer 1 frozen}         & False            & False               & False              &  \\ 
\textbf{layer 2 frozen}         & False            & False              & False               &  \\ 
\textbf{training iterations} & 51           & 51                 & 51                 &  \\ 
\textbf{no literals in summary} & False     & -                  & -                  &  \\ 
\bottomrule
\end{tabular}
\caption{Hyper-parameters of the three different R-GCN models}
\end{table}

\section{(K)-forward Bisimulation ({AIFB})}
\label{sec:aifbKB}
\begin{figure}[H]	
\centering
   \includegraphics[width=\linewidth]{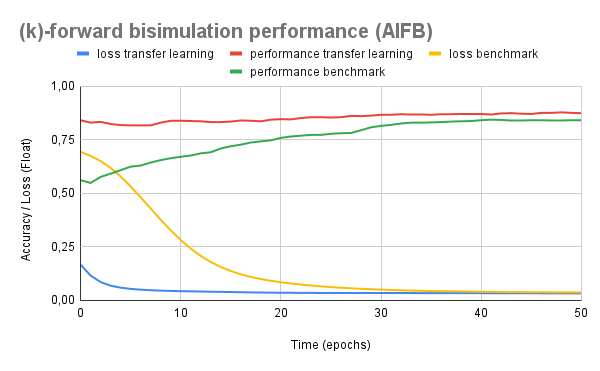}
   \caption{Graph showing the performance of the summarization relation (k)-forward bisimulation on the dataset AIFB}
   \label{fig:aifb_kb}

\end{figure}

\section{Attributes Summary and (K)-forward Bisimulation (MUTAG)}
\label{sec:mutagPCKB}
\begin{figure}[H]	
\centering
   \includegraphics[width=\linewidth]{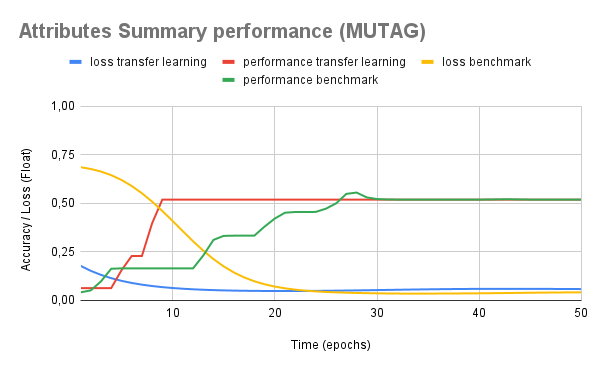}
   \caption{Graph showing the performance of the summarization relation PC on the dataset MUTAG}

\end{figure}

\begin{figure}[H]	
\centering
   \includegraphics[width=\linewidth]{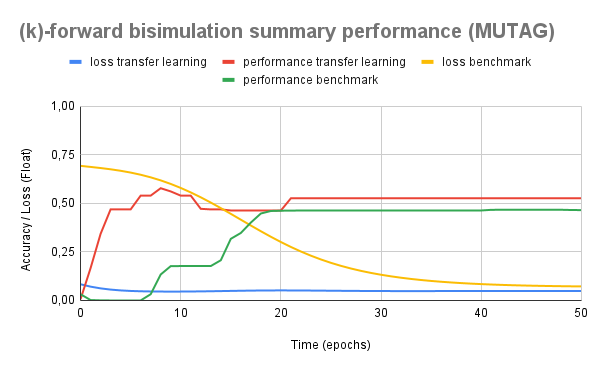}
   \caption{Graph showing the performance of the summarization relation (k)-forward bisimulation on the dataset MUTAG}

\end{figure}

\section{Attributes Summary and (K)-forward Bisimulation (AM)}
\label{sec:amPC}
\begin{figure}[H]	
\centering
   \includegraphics[width=\linewidth]{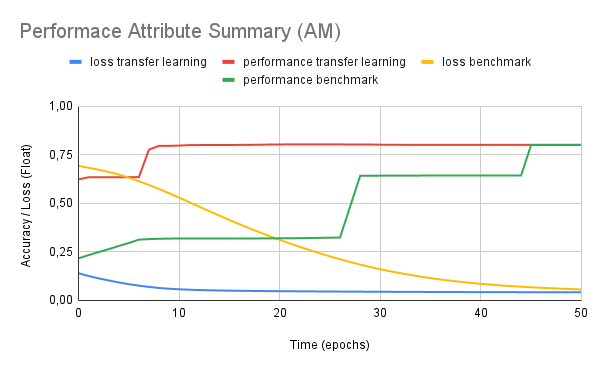}
   \caption{Graph showing the performance of the summarization relation PC on the dataset AM}

\end{figure}

\begin{figure}[H]	
\centering
   \includegraphics[width=\linewidth]{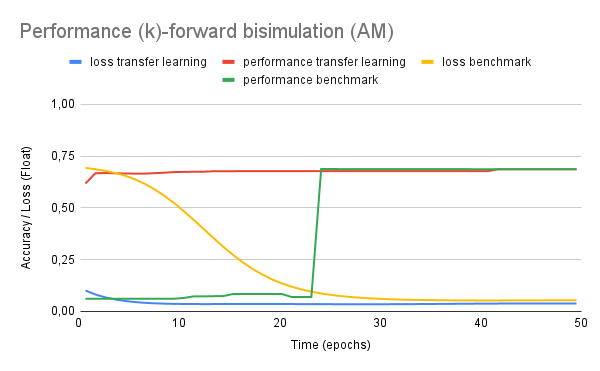}
   \caption{Graph showing the performance of the summarization relation (k)-forward bisimulation on the dataset AM}
    \label{fig:amKB}
\end{figure}

\section{The Effect of Leaving out Literals}
\label{sec:aifbPCnoLit}
\begin{figure}[H]	
\centering
   \includegraphics[width=\linewidth]{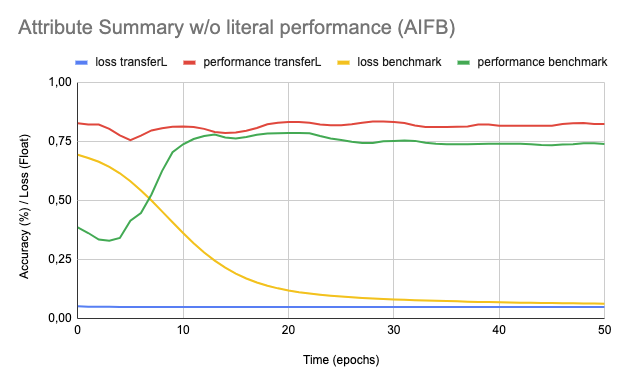}
   \caption{Graph showing the performance of the Attributes Summary without literals on the dataset AIFB.}

\end{figure}

\begin{figure}[H]	
\centering
   \includegraphics[width=\linewidth]{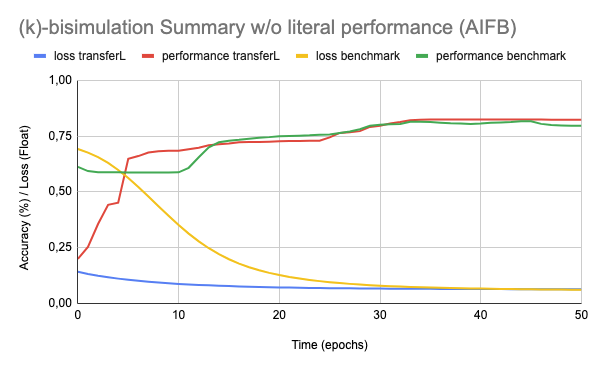}
   \caption{Graph showing the performance of the (k)-forward bisimulation Summary without literals on the dataset AIFB.}

\end{figure}

\balance

\begin{figure}[H]	
\centering
   \includegraphics[width=\linewidth]{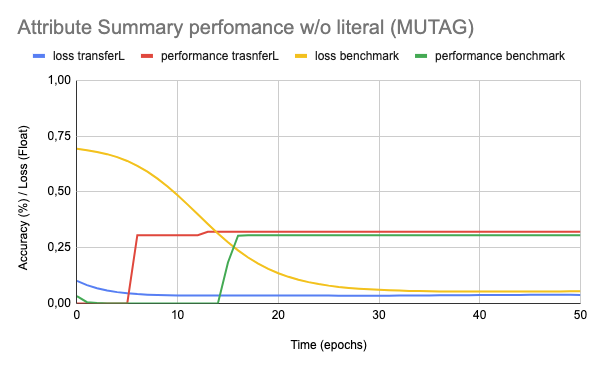}
   \caption{Graph showing the performance of the Attributes Summary without literals on the dataset MUTAG.}

\end{figure}

\begin{figure}[H]	
\centering
   \includegraphics[width=\linewidth]{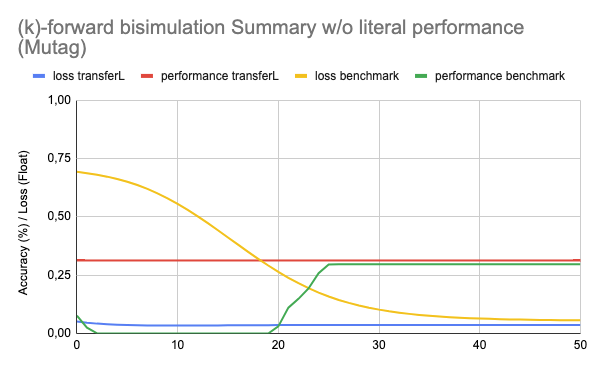}
   \caption{Graph showing the performance of the (k)-forward bisimulation Summary without literals on the dataset MUTAG.}

\end{figure}

\end{document}